\documentclass[journal]{IEEEtran}
\usepackage{amsmath,amssymb,amsfonts}
\usepackage{cite}
\usepackage{graphicx}
\usepackage{algorithm,algorithmic}
\usepackage{textcomp}
\usepackage[table]{xcolor}
\usepackage[caption=false,font=footnotesize]{subfig}
\usepackage{graphics}

\usepackage[inline]{enumitem}
\usepackage[hyphens]{url}
\usepackage{booktabs}

\graphicspath{{./fig/}}

\usepackage[hidelinks]{hyperref}

\definecolor{mygreen}{RGB}{255, 255, 255}

\DeclareMathOperator*{\argmin}{arg\,min}
\newcommand{\fref}[1]{Figure~\ref{#1}}
\newcommand{\sref}[1]{Section~\ref{#1}}
\newcommand{\eref}[1]{Equation~\ref{#1}}
\newcommand{\tref}[1]{Table~\ref{#1}}
\newcommand{\aref}[1]{Appendix~\ref{#1}}

\begin{document}
\title{Multimodal Diffeomorphic Registration with Neural ODEs and Structural Descriptors}
\author{Salvador Rodriguez-Sanz  and Monica Hernandez
\thanks{This work has been supported with the funding from Ministerio de Ciencia, Innovacion y Universidades Trust-BEyE PID2022-138703OB-I00, Gobierno de Aragon Orden ECU/1871/2023 PROY-B50-24, RICORS network of inflamatory diseases from Carlos III Health Institute Network RD24/0007/0022 and COS2MOS research group T64\_23R. Salvador Rodriguez-Sanz is also supported with Orden ECU/592/2024 Gobierno de Aragon predoctoral grant (2024-2028). The funders had no role in study design, data collection and analysis, decision to publish, or preparation of the manuscript.}
\thanks{Salvador Rodriguez-Sanz is with Aragon Institute for Engineering Research (I3A), University of Zaragoza (e-mail: srodsanz@unizar.es). }
\thanks{Monica Hernandez is with Aragon Institute for Engineering Research (I3A), University of Zaragoza (e-mail: mhg@unizar.es).}}

\maketitle

\begin{abstract}
This work proposes a multimodal diffeomorphic registration method using Neural Ordinary Differential Equations (Neural ODEs). Nonrigid registration algorithms exhibit tradeoffs between their accuracy, the computational complexity of their deformation model, and its proper regularization. In addition, they also assume intensity correlation in anatomically homologous regions of interest among image pairs, limiting their applicability to the monomodal setting. Unlike learning-based models, we propose an instance-specific framework that is not subject to high scan requirements for training and does not suffer performance degradation at inference time on modalities unseen during training. Our method exploits the potential of continuous-depth networks in the Neural ODE paradigm with structural descriptors, widely adopted as modality-agnostic metric models which exploit self-similarities on parameterized neighborhood geometries. We propose three different variants that integrate image-based or feature-based structural descriptors and nonstructural image similarities computed by local mutual information. We conduct extensive evaluations on different experiments formed by scan dataset combinations and show surpassing qualitative and quantitative results compared to state-of-the-art baselines adequate for large or small deformations, and specific of multimodal registration. Lastly, we also demonstrate the underlying robustness of the proposed framework to varying levels of explicit regularization while maintaining low error, its suitability for registration at varying scales, and its efficiency with respect to other methods targeted to large-deformation registration.
\end{abstract}


\section{Introduction}
\label{sec:introduction}

\IEEEPARstart{D}{iffeomorphic} registration is a crucial task in medical image analysis that aims to estimate a dense spatial transformation that aligns two given images, so this transformation is smooth and has a smooth inverse.
Deformations in medical images can arise from disease, growth, or motion, and registration algorithms compensate for these deformations to analyze and quantify anatomical variability. The interest of medical imaging in requiring these transformations to be global diffeomorphisms is motivated by some of their key features: (1) The conservation of shape topology; (2) No introduction of sharp artifacts or collapsing voxels; (3) Invertibility with the same degree of smoothness. These properties are necessary for proper usability of the registration methods \cite{toga2001roleregistration}, as a registration algorithm with these features allows the definition of a normalized coordinate system whose versatility facilitates other related applications, such as the detection of regions of interest by template-based segmentation methods \cite{ashburner2005unified} or atlas building \cite{joshi2004unbiaseddiffeomorphic}. Diffeomorphisms have traditionally been estimated in the large deformation paradigm by time-dependent variational optimization \cite{beg2005lddmm}, geodesic shooting \cite{miller2006geodesic}, or stationary velocity fields \cite{hernandez2009svf, arsigny2006logeuclidean}.

The main limitations of registration algorithms are traditionally well-grounded by several factors: (1) The non-convexity of the registration problem, so numerical optimization methods converge to local minima; (2) The underlying optimization incurs high computational complexity, given by time-discrete first-order or second-order methods; (3) The presence of domain shifts, which induces method bias. Although the extensive literature on registration addresses items (1) or (2), domain shifts in (3) range over heterogeneous limitations like intensity inhomogeneities, bias field, or different spatial resolution, each of them requiring particular preprocessing.

In this work, we focus on the domain shifts that result from modality shifts.  Modality-agnostic registration induces an additional complexity gap in addition to the geometry shift present in the monomodal setting. However, efficient approaches for multimodal registration would benefit clinical applications that require the integration of different sources, such as: (1) Morphology comparison of tissues of different density, for example, by Magnetic Resonance Imaging (MRI) and Computed Tomography (CT); (2) Surgical planning or image-guided intervention (e.g., in applications requiring deformable 2D-3D registration \cite{uneri2016deformable2d3d}); (3) Anatomical surface reconstruction by combining features from multi-contrast sequences (e.g., cortical reconstruction \cite{gopinath2025reconallclinical}).

However, current approaches for multimodal registration are challenged by significant limitations: (1) High variability, since there are many different modalities available for single patient examinations; (2) Intrinsic uncorrelatedness in intensity or local feature patterns across modalities, so numerical optimization methods fail to capture local anatomical similarity with common similarity metrics; (3) Lack of defining features. Due to these, multimodal registration is commonly restricted to specific domain shifts \cite{lu2022plospanacea}, fused by a domain transfer method to a monomodal setting \cite{iglesias2023multimodal}, or reduced to landmark methods \cite{wang2023robustkeypoint}.

Despite its challenges, multimodal registration was first tackled two decades ago by information-theoretic measures such as mutual information \cite{wells1996mutualinformation, viola1997allignment, gholipour2009information} or the correlation ratio \cite{roche1998correlationratio}. Mutual information has been successful in low-rank parametric methods, such as rigid registration \cite{pluim2003mutualinformation} or free-form deformations \cite{rueckert2006diffeo,sideri2022multimodal}. Global, local or normalized mutual information \cite{studholme1999multimodal} lacks structural intensity information, so other structural representations have been proposed, such as feature-based encoding with self-similarity descriptors \cite{heinrich2012mind}, hierarchical attributes \cite{shen2001hammer}, physics-inspired losses for robust local comparison in ultrasound \cite{wein2013globalregistrationlc2}, normalized gradient fields \cite{haber2006intensity} or registration by warping in feature space and proximal splitting methods \cite{steinbrucker2009largedeformation, heinrich2014convex, oquab2024dinov2}. The high variability in different medical imaging modalities influences different registration approaches to specifically work conditioned to concrete anatomies, combinations of modalities, or segmentation features, among other priors. Within these limits, we propose a multimodal registration framework in brain scans and in the large deformation paradigm, working exclusively on image intensity space.

\subsection{Statement of Contribution}
This work proposes a pairwise strategy to address multimodal diffeomorphic registration on domain shifts between T1w-T2w brain scans via Neural Ordinary Differential Equations (Neural ODEs). First, we study the potential of structural representations in the achievement of accurate registration in comparison with information-theoretic or segmentation-overlap losses. We work exclusively on intensity-based dissimilarity models and perform extensive evaluation in challenging brain scan benchmarks determined by varying degrees of atrophy. The freedom of choosing different metric models enables us to compare our proposed approach in three variants: (1) a structural descriptor based on sum-of-squared differences on image space; (2) a dense descriptor trained by contrastive learning and computed on feature space; and (3) local mutual information registration. Although these variants achieve promising results, model (1) significantly outperforms all state-of-the-art baselines in most of our experiments.
\section{Related work}
\label{sec:related-work}
The large deformation paradigm was first introduced in the Large Deformation Diffeomorphic Metric Mapping (LDDMM) method \cite{beg2005lddmm}, which provides a variational framework for numerical optimization of dense displacements in time-dependent velocity fields defined in a Reproducing Kernel Hilbert Space (RKHS). Further work has proposed variants with improved features on the registration output or its algorithmic efficiency, such as the stationary parametrization \cite{hernandez2009svf, arsigny2006logeuclidean}, symmetric registration \cite{avants2008syn}, band-limited spectral methods for geodesic shooting \cite{zhang2018bl} or PDE-constrained optimization \cite{hernandez2019bandlimitedstokes, mang2016h1regularizers} useful for obtaining richer physics-inspired models. More recent advances have focused on the integration of neural network architectures for the diffeomorphic registration task, mainly using stationary velocity fields, with VoxelMorph the main baseline \cite{balakrishnan2019voxelmorph}. Subsequent work has gradually integrated visual transformers (ViTs) in TransMorph \cite{chen2022transmorph}, diffusion-based generation of velocity fields \cite{kim2022diffusemorph, starck2025diffdef}, or implicit neural representations \cite{tian2025nephi, shanlin2024regneuralfields, han2023diffeovelocityfields}. Other approaches fundamentally incorporate new features in registration networks to tackle some of the challenges inherent in their numerical optimization, such as cycle consistency \cite{boah2021cyclemorph}, coarse-to-fine multiresolution training \cite{mok2020lapirn, mok2021conditionalreg} or equivariance \cite{greer2025carl}. All these methods are learning-based and register fast at inference time \cite{hernandez2024insights}, being surpassed in the pairwise registration paradigm by Neural Ordinary Differential Equations (Neural ODEs) \cite{yifan2022nodeo, hernandez2025pdelddmmnodeo}. Despite this superiority, most of these methods are not specifically designed to be robust in different domain shifts that occur by different modality combinations.

In this multimodal setting, there is intensity uncorrelatedness among corresponding anatomical regions of interest. Novel methods have achieved accurate results under certain assumptions. State-of-the-art learning-based methods have proposed reducing the problem by image-to-image translation \cite{iglesias2023multimodal,lu2022plospanacea, demir2024multimodalsegmentations}, learning parametric free-form deformations on B-spline basis functions \cite{rueckert2006diffeo, sideri2024sinr}, or dictionary learning \cite{cao2014multimodalmicroscopysparsecoding}. The challenge of modeling modality-agnostic similarity functions has been addressed by self-similarity descriptors that encode local patterns \cite{heinrich2012mind, shechtman2007localselfsimilarity, wachinger2012entropy, shen2014multimodalmultispectral} or gradient field orientation \cite{haber2006intensity}. Other advances have used finite-dimensional decompositions on scale-space basis with encoder-based metric learning \cite{xu2021multiscale},  sign-agnostic cross correlation and gradient consistency \cite{demir2024multigradicon}, trained differentiable approximations by supervised metrics or contrastive learning \cite{ronchetti2023disa, mok2024modalityagnostic}, neural optimal transport \cite{kim2025otmorph}, or directly fused the problem to the segmentation domain \cite{hoffmann2022synthmorph, chen2024transmatch}. Recent work has adapted proximal splitting methods to warp in feature space by \texttt{MIND} \cite{heinrich2012mind} or \texttt{DINOv2} \cite{oquab2024dinov2} features \cite{heinrich2014convex, siebert2025convexadam, song2024dinoreg, song2025dinoreg, steinbrucker2009largedeformation}. Although self-supervised features are in active development for general-purpose semantic or visual tasks \cite{simeoni2025dinov3}, it is still a very challenging goal to model accurate, robust to noise, efficient and dense similarity functions across domain shifts for medical image registration.

\section{Methodology}

\subsection{Diffeomorphic Registration}

Let $I_{0}$ and $I_{1}$ be two images (functions) $I_{k}: \Omega \rightarrow \mathbb{R}$ where $\Omega \subseteq \mathbb{R}^{d}$ is a closed and bounded domain, $d \in \{2, 3\}$ and each $I_{k} \in L^{2}(\Omega)$. Diffeomorphic registration estimates a diffeomorphism $\varphi^{\ast}$ that warps $I_{0}$ to $I_{1}$ by a variational optimization problem

\begin{equation}
\label{eq:variational-registration}
\varphi^{\ast} := \argmin_{\varphi \in \text{Diff}(\Omega)} \mathcal{S}(I_{0} \circ \varphi^{-1}, I_{1}) + \mathcal{R}[\varphi].
\end{equation}

The set $\text{Diff}(\Omega)$ is the manifold of diffeomorphisms in $\Omega$, the term $\mathcal{S}(\cdot, \cdot)$ acts as a dissimilarity term and $\mathcal{R}[\cdot]$ is a regularizer acting as a penalty over the transformation cost, so the registration problem minimizes an energy. This problem is infinite-dimensional and highly nonconvex in $\varphi$, so its regularization is unavoidable, and the resulting optimization is made tractable by numerical methods. 

Common choices of the similarity function $\mathcal{S}(\cdot, \cdot)$ for monomodal registration are the norm $L^{2}$ \cite{beg2005lddmm}, normalized cross correlation \cite{avants2008ncc} or mutual information \cite{wells1996mutualinformation}. Regarding the regularizer $\mathcal{R}[\cdot]$, most decisions consider diffusion regularization \cite{vercauteren2008symmetric}, cycle composition consistency \cite{ greer2021icon}, spatially varying kernels \cite{niethammer2019metriclearning} or hyperelastic models \cite{burguer2013hyperelasticregularizer}.

In this work, we propose a framework for diffeomorphic registration and compare different options for $\mathcal{S}$, considering a fixed regularization scheme $\mathcal{R}$. 

\subsubsection{Pairwise Registration by velocity fields}
\label{subsec:pairwise-node}
The \textsl{Pairwise} diffeomorphic registration problem aims to estimate the best diffeomorphism $\varphi^{\ast}$ according to the variational problem from \eref{eq:variational-registration} conditioned to a pair of fixed $I_{0}$ and moving $I_{1}$ images. The deformation between $I_{0}$ and $I_{1}$ is parameterized by a flow $\Phi: [0, 1] \times \Omega \rightarrow \Omega$. Concretely, $\Phi$ is the ODE flow of the transport initial value problem \cite{chen2011imagesequence}

\begin{equation}
\label{eq:transport-equation}
\begin{cases}
\partial_{t} \Phi(x, t) = v(\Phi(x, t)), \\
\Phi(x, 0) = x.
\end{cases}
\end{equation}

Here, the velocity field $v$ is solved in $L^{2}([0, 1], V)$, where $V$ is an RKHS obtained from a semi-definite positive operator $\mathcal{K}$. The initial value problem in \eref{eq:transport-equation} is stationary, so the target $\varphi$ is obtained by the group exponential map in $v$, that is, $\varphi = \text{Exp}(v)$. The optimization problem which relates the pairwise registration in the Eulerian frame $v$ reads

\begin{equation}
\label{eq:kkt-registration}
\begin{aligned}
\min_{v \in L^{2}([0, 1])} \quad & \mathcal{J}[\varphi] = \mathcal{S}(I_{0} \circ \varphi^{-1}, I_{1}) + \mathcal{R}[\varphi], \\
\text{subject to} \quad & \partial_{t} \Phi = v \circ \Phi \hspace{15px} \text{ in } \Omega \times (0, 1], \\
                        & \Phi = \text{id}_{\Omega} \hspace{35px} \text{in } \Omega \times \{0\}.
\end{aligned}
\end{equation}

We follow an optimize-then-discretize approach \cite{haber2010numericaloptimization, kidger2021node}. This constrained problem is nonlinear with respect to the flow $\Phi$. We form its augmented Lagrangian $\mathcal{L}$, analogously to \cite{zhang2019anodev2}, by
\begin{equation}
\mathcal{L} =  \mathcal{J}[\varphi] + \int_{0}^{1} \langle \lambda(t), \partial_{t} \Phi - v \circ \Phi \rangle \, \text{d}t + \langle \mu, \Phi_{0} - \text{id}_{\Omega} \rangle.
\end{equation}

The bilinear map $\langle \cdot, \cdot \rangle$ is the inner product in $L^{2}(\Omega)$, and $\Phi_{t} \equiv \Phi(\cdot, t)$. The terms $\lambda: [0, 1] \times \Omega \rightarrow \Omega$ and $\mu: \Omega \rightarrow \Omega$ are the corresponding adjoints or Lagrangian multipliers for each constraint. The first-order optimality conditions on the Karush-Kuhn-Tucker (KKT) system impose the Gâteaux derivative $\delta{\mathcal{L}}(\Phi; \delta\Phi)$ to vanish 
\begin{equation}
\delta{\mathcal{L}}(\Phi; \delta\Phi) = \dfrac{d}{d\varepsilon} \mathcal{L}[\Phi + \varepsilon \delta \Phi] \Bigg |_{\varepsilon=0} = 0.
\end{equation}
By this condition, the dynamics of the adjoint $\lambda$ is
\begin{equation}
\label{eq:adjoint-equation}
\begin{cases}
\partial_{t} \lambda = -  \lambda \partial_{x} v(\Phi(x, t)), \\
\lambda(1) = - \dfrac{\delta \mathcal{J}}{\delta \varphi}.
\end{cases}
\end{equation}

The term $\frac{\delta \mathcal{J}}{\delta \varphi}$ stands for the functional derivative of $\mathcal{J}$ with respect to $\varphi$. These equations correspond to the stationary case and are well-known in the literature, for example, in \cite{vialard2011diffeoadjoint}. 

\subsubsection{Neural Ordinary Differential Equations (Neural ODEs)}

We encode the velocity field $v_{\theta}$ as a neural network with trainable parameters $\theta$ in a Neural ODE \cite{yifan2022nodeo, chen2018node}. The KKT system in condition $\partial\mathcal{L}/\partial \theta = 0$ relates the gradients of the loss term $\mathcal{J}$ with respect to the trainable parameters by

\begin{equation}
\label{eq:loss-parameters-derivative}
\dfrac{\partial \mathcal{J}}{\partial \theta}[\varphi] = \int_{0}^{1} \left\langle \lambda(t), \dfrac{\partial (v_{\theta} \circ \Phi)}{\partial \theta} \right\rangle \, \text{d}t.
\end{equation}

The network output is $\varphi$. The initial value problem of \eref{eq:adjoint-equation} can be integrated to compute the adjoint $\lambda$ using an ODE solver. Once the adjoint is computed, the gradients of the loss term $\mathcal{J}$ with respect to the trainable parameters are computed by \eref{eq:loss-parameters-derivative}. The final map $\varphi = \Phi_{1}$ is given by an ODE solver between time instants $t_{0} = 0$ and $t_{1} = 1$ and step size $h$ by

\begin{equation}
\Phi_{t_{1}} = \text{ODESolver}(\Phi_{t_{0}}, v_{\theta}, t_{0}, t_{1}, h).
\end{equation}

This velocity field $v_{\theta}$ is given by a neural network, and the problem is well-posed. The resulting Neural ODE is trained with the following loss, sharing the same notation as \eref{eq:variational-registration}: 

\begin{equation} 
\label{eq:node-optimization-target}
\mathcal{L}_{\text{NODE}} = \mathcal{S}(I_{0} \circ \varphi^{\ast}, I_{1}) + \lambda_{J} \mathcal{L}_{J} + \lambda_{\text{grad}} \mathcal{L}_{\text{grad}} + \lambda_{\text{mag}}\mathcal{L}_{\text{mag}}.  
\end{equation}

The remaining terms are regularizers that encourage desirable local properties in $\varphi$. In this work, these regularizers are \cite{hernandez2025pdelddmmnodeo}
\begin{equation}
\mathcal{L}_{J}[\varphi] = \int_{\Omega} \max(0, - \det(\text{Jac}[\varphi](\mathbf{x}) + \varepsilon) \, \text{d}\mathbf{x},
\end{equation}
being $0 < \varepsilon < 1$ a penalty-barrier hyperparameter and 
\begin{equation}
\mathcal{L}_{\text{grad}}[\varphi] = \int_{\Omega} \lVert \nabla\varphi(\mathbf{x}) \rVert^{2}_{2} \, \text{d}\mathbf{x}.
\end{equation}

The third term on the loss functional from \eref{eq:node-optimization-target} is given by the $V$-regularizer
\begin{equation}
\mathcal{L}_{\text{mag}}[\varphi] = \int_{0}^{1} \lVert v_{\theta}(t, \Phi(t, x)) \rVert^{2}_{V} \, \text{d}t, \hspace{5px} \lVert v \rVert_{V}^{2} \equiv \lVert
\mathcal{K} v \rVert_{L^{2}}^{2}.
\end{equation}

These regularizers determine the deformation model learned by the neural network in the velocity field $v_{\theta}$. This problem can be extended with geometrical constraints to impose more realistic models in the final map $\varphi$ \cite{hernandez2025pdelddmmnodeo, hernandez2021pdelddmm}, inspired by continuum mechanics.

\begin{figure*}[t]
    \centering
    \includegraphics[width=0.85\textwidth]{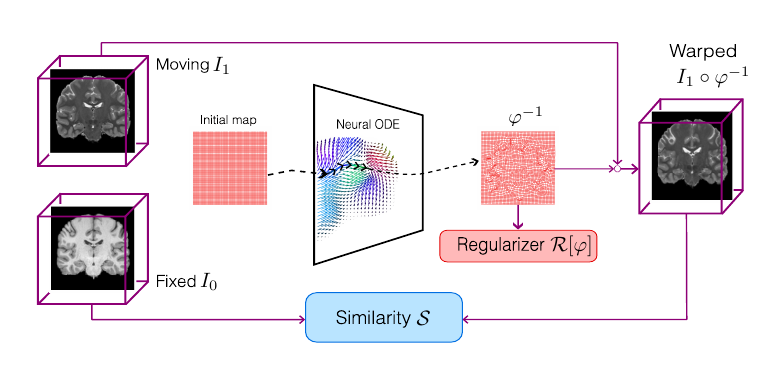}
    \caption{\textbf{Overview of the registration framework}. We represent our adopted pairwise registration method for our tridimensional multimodal setting. For a given pair of images, the Neural ODE backbone is trained to compute the domain transformation on $\varphi$, optimized by a modality-agnostic similarity $\mathcal{S}$.}
    \label{fig:multimodal-framework}
\end{figure*}

\subsection{Structural Descriptors}
\label{subsec:structural-descriptors}

For multimodal registration, local features that are agnostic to domain shifts depend on intensity, gradient orientation, edges, and texture. Such representations constitute dense descriptors that have traditionally been based on exploiting self-similarity and were first introduced in \cite{shechtman2007localselfsimilarity} to compute fine-grain representations of visual entities in images or videos and address tasks such as template-based matching.

Given an image $I$ in $\mathbb{R}^{d}$ as input, for each voxel $\mathbf{x} \in \Omega$, self-similarity descriptors compute pairwise distances in a spatial search region $R$. The region $R$ in this case contains two symmetrical 6-neighborhoods, so $|R| = 12$. For each of the sampled positions, a distance-based representative of self-similarities is computed by the expression

\begin{equation}
\label{eq:descriptor-ss-token}
\text{D}(I, \mathbf{x}, \mathbf{r}) = \exp \left(-\dfrac{d_{P}(I, \mathbf{x}, \mathbf{x}+\mathbf{r})}{\text{Var}(I, \mathbf{x})} \right), \hspace{10px} \mathbf{r} \in R.
\end{equation}

The term $d_{P}$ is a given distance over elements in a patch and $\text{Var}(I, \textbf{x})$ is an estimate of local variance. This computation is composed of an exponential filter acting as a low-band pass filter, so the resulting token suppresses the influence of high-frequency noise. In the computation of $\text{D}$, the regions $R$ and $P$ remain fixed and are hyperparameters of the spatial search scheme. 

\subsubsection{Modality Independent Neighborhood Descriptor (MIND)}
\label{subsec:mind-descriptor}

A widely used descriptor based on self-similarity is the \texttt{MIND} descriptor \cite{heinrich2012mind}. From the notation in \eref{eq:descriptor-ss-token}, the \texttt{MIND} descriptor computes the patch distance $D_{P}$ as the Mean Squared Error (MSE) of related positions in a neighborhood region, by
\begin{equation}
d_{P}(I, \mathbf{x}_{1}, \mathbf{x}_{2}) = \sum_{p \in P} (I(\mathbf{x_{1}} + p) - I(\mathbf{x}_{2} + p))^{2}.
\end{equation}

This cumulative distance $d_{P}$ is computed by a convolution kernel. The region $R$ remains fixed and the spatial search of the region $P$ is parametrized by a radius $r$ and a dilation $d_{r}$.  The distance $d_{P}$ contains a sum of $(2r+1)^{d}$ terms and is computed $|R|$ times for each voxel $\mathbf{x} \in \Omega$. The variance measure $\text{Var}(I, \mathbf{x})$ for each image $I$ at each voxel location $\mathbf{x}$ is estimated locally 

\begin{equation}
\text{Var}(I, \mathbf{x}) = \dfrac{1}{6} \sum_{p \in P} d_{P}(I, \mathbf{x}, \mathbf{x}+p).
\end{equation}

The choice of radius $r$ or dilation factor $d_{r}$ in the convolution conditions features attributes on varying scales and is tuneable for the descriptor computation. Image similarity becomes the Mean Squared Error (MSE) of the descriptor differences over voxel locations.

This descriptor is a baseline to obtain modality-agnostic features in medical registration, especially in state-of-the-art methods such as \cite{ demir2024multigradicon, siebert2025convexadam, heinrich2013mrfdeeds} or seminal work \cite{heinrich2014convex, heinrich2015multimodaldeedsbcv}.

\subsubsection{Differentiable Approximation by Contrastive Learning}
\label{sec:differentiable-approximation-contrastive}
We explore other possibilities to encode the descriptor $\text{D}$ from \eref{eq:descriptor-ss-token} with metric learning. Concretely, given the input image $I$, we encode the dense descriptor with \cite{mok2024modalityagnostic} by a neural network $F_{\theta}$. For each image $I$, we augment $I$ by sampling $n+1$ control points satisfying $P_{0} \leq P_{1} \leq \cdots \leq P_{n}$ and form an intensity transform map given by the Bézier curve
\begin{equation}
B(t) =  \sum_{i=0}^{n} P_{i} \, b_{i, n}(x), \hspace{10px} t \in [0, 1],
\end{equation}
where each $b_{i, n}$ is the $i-$th Bernstein polynomial of degree $n$, i.e.,
\begin{equation}
b_{i, n}(t) = \begin{pmatrix}
n \\
i \\
\end{pmatrix}
t^{i}(1-t)^{n-i}.
\end{equation}
Since the control points $P_{i}$ are arranged so that they are an increasing sequence, this transformation is monotone and, therefore, one-to-one. Indeed, the derivative is
\begin{equation}
B'(t) = n \sum_{i=0}^{n-1} (P_{i+1} - P_{i})b_{i, n-1}(t) \geq 0, \hspace{10px} t \in [0, 1].
\end{equation}
For contrastive training, we stochastically augment the input image $I$ to another image $I'$.  With a fixed probability $p$, we compute $I'$ by applying the Bézier map $B$ to the image $I_{2}(\mathbf{x}) = 1 - I(\mathbf{x})$ with probability $p$ or to $I$ with probability ${1-p}$, in order to reproduce possible domain shifts during training \cite{mok2024modalityagnostic, pielawski2020comir}. If $I$ has spatial resolution $H \times W \times D$, the network output $F_{\theta}(I)$ computes a feature map $H \times W \times D$ on which we form pixel-wise tokens via \eref{eq:descriptor-ss-token} where $d_{P}$ is set to be the sum-of-squared differences. 

The network $F_{\theta}$ is trained by computing in every training iteration the feature tokens of $I$ and $I'$ and obtain normalized tokens $d_{I}(\mathbf{x}) / \lVert d_{I}(\mathbf{x})\rVert_{2}$ or $d_{I'}(\mathbf{x}) / \lVert d_{I'}(\mathbf{x}) \rVert_{2}$, respectively, each of which has $24$ components, since we compute the self-similarity from \eref{eq:descriptor-ss-token} for two dilation factors ${d_{r} \in \{1, 2\}}$ and stack these descriptors. Network training is performed by sampling $N_{k}$ locations, and positive pairs are determined by identical anatomical locations determined by voxel positions. This criterion is then applied in the voxel-level contrastive loss:
\begin{equation}
\label{eq:contrastive-token-loss}
\scalebox{0.81}{$
\ell(\mathbf{x}_{i}) = 
\log \dfrac{\exp(d_{I}^{T}(\mathbf{x}_{i})d_{I'}(\mathbf{x}_{i})/\tau )}{\exp(d_{I}^{T}(\mathbf{x}_{i})d_{I'}(\mathbf{x}_{i})/\tau ) + \sum_{j \neq i} \exp(d_{I}^{T}(\mathbf{x}_{i}) d_{I'}(\mathbf{x}_{j})/\tau)},
$}
\end{equation}
where $\tau$ is the temperature parameter, and maximization encourages the descriptor of non-matching locations to be orthogonal. This criterion is suitable for dense registration, as input pairs are aligned by an affine registration algorithm to an atlas in our setting. However, other choices are common depending on the visual task, such as patch-distance thresholding \cite{xie2021propagateyourself}. The total loss of minimization by training is then given by

\begin{equation}
\mathcal{L}_{\text{Descriptor}} = - \dfrac{1}{N_{k}} \sum_{i=1}^{N_{k}} \ell(\mathbf{x}_{i}).
\end{equation}

Registration is achieved by minimizing the dissimilarity of the descriptors \cite{ronchetti2023disa, mok2024modalityagnostic}. With the notation of \eref{eq:node-optimization-target}, the dissimilarity is
\begin{equation}
\mathcal{S}(I_{0}, I_{1}) = 1 - \int_{\Omega} \left\langle 
\dfrac{d_{I_{0}}(\mathbf{x})}{\lVert d_{I_{0}}(\mathbf{x}) \rVert}, \dfrac{d_{I_{1}}(\mathbf{x})}{\lVert d_{I_{1}}(\mathbf{x}) \rVert}
\right\rangle \, d\mathbf{x}.
\end{equation}
We consider that registration is performed in a two-stage training, and the network $F_{\theta}$ is already trained and frozen at registration time for each registration pair. In this case, $\text{Var}(I, \mathbf{x})$ is estimated for each $\mathbf{x} \in \Omega$ as in \sref{subsec:mind-descriptor}.

\subsection{Local Mutual Information}
Consider $I_{0}$ and $I_{1}$ as in \eref{eq:kkt-registration}. We divide each image $I_{i}$ into $n_{p}$ patches and denote by $I_{i}^{j}$ the patch $j$, $j \in \{1, \ldots, n_{p} \}$. Then, we compute local mutual information by
\begin{equation}
\mathcal{I}(I_{0}, I_{1}) = \dfrac{1}{n_{p}}\sum_{i=1}^{n_{p}} p(I_{0}^{i}, I_{1}^{i}) \log \dfrac{p(I_{0}^{i}, I_{1}^{i})}{p(I_{0}^{i}) p(I_{1}^{i})}.
\end{equation}
Each term $p(I, J)$ stands for the joint probability density function of images $I$ or $J$ estimated using a differentiable kernel density method. The registration by our chosen Neural ODE maximizes $\mathcal{I}$ by
\begin{equation}
\mathcal{S}(I_{0}, I_{1}) = - \mathcal{I}(I_{0}, I_{1}),
\end{equation}
with $\mathcal{S}$ the same as \eref{eq:kkt-registration}. This metric is nonstructural, but has shown good registration results for low-rank affine or B-spline models, as well as nonparametric multimodal registration for T1w-T2w pairs.

\subsection{Implementation Details}

\subsubsection{Neural ODE}

We used Euler solver to integrate forward in time and backward for the corresponding adjoint given its empirical performance, as other solvers with fixed step size increase the number of evaluations. We selected the step size of the solver as $h = 0.005$ for all our experiments and a learning rate of $5 \cdot 10^{-3}$.

\begin{figure}[t]
    \centering
    \def\svgwidth{\linewidth}
    \includegraphics[width=0.5\textwidth]{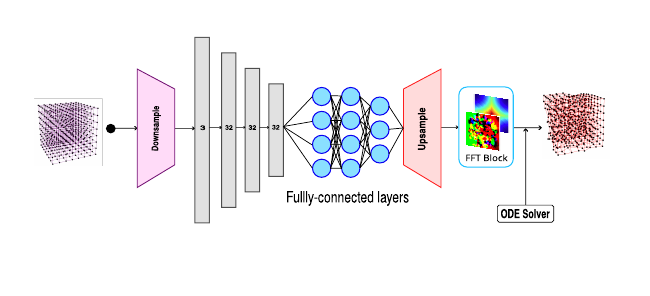}
    \caption{\textbf{Overview of the Neural ODE backbone}. The input is the sampled image domain $\Omega$ and it is forwarded to a downsampling layer, followed by convolutions and two linear layers which result in the velocity field $v_{\theta}$.}
    \label{fig:overview-node-network}
\end{figure}

We choose $\mathcal{K}$ as the semi-definite positive Laplacian operator $\mathcal{K} = (\gamma  - \alpha \Delta)^{s}$ \cite{hernandez2025pdelddmmnodeo}, being $\gamma=1$ and $\alpha = 5 \cdot 10^{-4}$, so it induces a RKHS $V$ over $L^{2}([0, 1])$. The convolution of this operator $\mathcal{K}$ is performed in the frequency domain.  Since this operator is self-adjoint as we assume velocity fields are Dirichlet homogeneous in the boundary of $\Omega$, and the Fourier transform is a $L^{2}$-isometry, we compute in the frequency domain
\begin{equation}
\langle v, w \rangle_{V} = \langle \mathcal{K}^{\ast} \mathcal{K}v, w \rangle_{L^{2}} = \langle \hat{\mathcal{K}} \cdot \hat{\mathcal{K}} \cdot \hat{v}, \hat{w} \rangle_{L^{2}}.
\end{equation}
The neural network contains a final layer that outputs $(\mathcal{K}^{\ast} \mathcal{K})^{-1} v$, and optimizes the registration energy from \eref{eq:kkt-registration}. The hyperparameters controlling the losses penalties \eref{eq:node-optimization-target} are set to $\lambda_{J} = 2.5$, $\lambda_{\text{grad}} = 5 \cdot 10^{-2}$, $\lambda_{\text{mag}} = 5 \cdot 10^{-5}$ and $\varepsilon = 0.1$ from \eref{eq:node-optimization-target}.  We select these parameters to leverage existing tradeoffs between registration accuracy and smoothness, so the ratios of negative Jacobians have the same order as the rest of the baselines, roughly a power of $10^{-2}$ on percentage scale. The network is trained in every pair with 300 epochs and performs one step of downsampling and upsampling of factor 2 in each axis. This network is represented in \fref{fig:overview-node-network}. 

\subsubsection{MIND Descriptor}
In this work, we consider descriptors computed with dilation factor $d_{r} = 2$ and radius or a receptive field $r = 1$. The combination of these values has been found to perform best over other choices. 

\subsubsection{Descriptor Network}
We used to represent $F_{\theta}$ a UNet-style encoder-decoder network with skip-connections per resolution level and two initial and final convolutional blocks. It has three downsampling and upsampling layers, and outputs a feature map of the same shape as the input image. We choose the temperature parameter $\tau$ from \eref{eq:contrastive-token-loss} as 0.05, and we compute the deep descriptors by stacking two tokens computed by \eref{eq:descriptor-ss-token} with two dilation factors of $d$, being $d \in \{ 1, 2 \}$. The network is trained at a learning rate $10^{-4}$ with batch size 1, the sampling number of anatomical positions to train on \eref{eq:contrastive-token-loss} is $N_{k} = 8196$, the number of control points in the intensity transformation is $n = 3$ and the probability of intensity shift on $I$ or $I_{2}$ to $I'$ is set to $p = 0.5$. The resulting dense descriptors have length 24. 

\begin{figure}[t]
    \centering
    \includegraphics[width=0.46\textwidth]{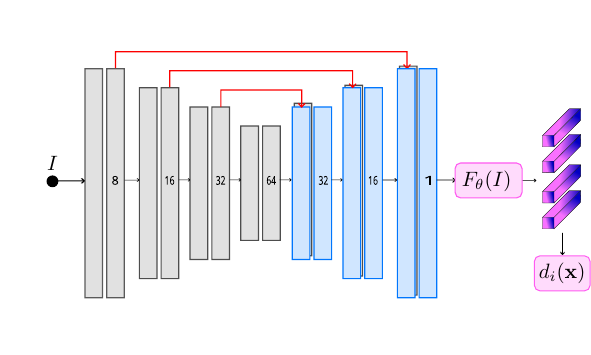}
    \caption{\textbf{Overview of the descriptor network $F_{\theta}$}.  The output channels of each of the convolutional blocks are represented for each stride level, and the resulting feature map $F_{\theta}$ is used to compute dense descriptors by \eref{eq:descriptor-ss-token}. We use bilinear upsampling on each decoder block.}
    \label{fig:overview-descriptor-network}
\end{figure}

We train a single descriptor network with our training partition of the OASIS-3 dataset described in \sref{subsec:datasets} and run all experiments with these weights. There is no coincidence in the training samples and the registration pairs selected for evaluation. An overview of the structure of the descriptor network is given in \fref{fig:overview-descriptor-network}.

\subsubsection{Local Mutual Information}
We discretize the measure into 16 bins and consider patches of size $n_{p} = 21$.

\section{Experiments}
\label{sec:results-discussion}

\subsection{Datasets}
\label{subsec:datasets}

The purpose of our method is to achieve accurate registration in large deformation registration, so we select our evaluation datasets accordingly. In particular, we choose available scans as fixed images from patients with different degrees of Alzheimer disease (AD), which is correlated with brain atrophy:

\begin{enumerate}[label=\alph*),
    listparindent=0pt,
    leftmargin=0pt,
    itemindent=2\parindent,
    parsep=1pt,
    topsep=2pt,
    partopsep=0pt]
    \item \textsl{OASIS-3}: We gather 3D brain MRI scans from Open Access Series Imaging Studies (OASIS-3) \cite{lamontagne2019oasis3}. OASIS-3 includes T1w and T2w sequences acquired at 1.5T and $\sim$1mm of isotropic resolution in a longitudinal study including healthy, suspected dementia, mild cognitive impairment and advanced stages of AD. We use for this work a population of 1018 scans.
    \item \textsl{IXI}: These T1w and T2w scans are acquired at 1.5T and $\sim$1mm isotropic resolution of only healthy subjects \cite{ixidataset}. We do not discard any scan by hospital origin. After manual discard of bad quality scans, our population contains 573 scans.
\end{enumerate}

\subsubsection{Dataset processing}
The datasets were preprocessed from scratch with Freesurfer to perform intensity normalization and bias field correction. Skull-stripping on these datasets was performed with \texttt{SynthStrip} \cite{hoopes2022synthstrip}. We manually cropped the multimodal ICBM152 atlas \cite{mazziotta2001atlas} to an affine space of $160 \times 224 \times 192$ and registered all our samples by an affine map to this space with $\sim$1mm isotropic resolution. Segmentation maps for evaluation are obtained with \texttt{SynthSeg} \cite{billot2023synthseg} on the aligned scans. Coronal views of each of the chosen fixed images can be seen in \fref{fig:atlases-coronal-views}.

\begin{table}
\caption{Test registration sets compiled from OASIS-3 and IXI for our experiments.}
\label{table:test-registration-sets}
\begin{tabular}{@{}llll@{}}
\midrule
\multicolumn{1}{l}{\textbf{Experiment Name}} & 
\multicolumn{1}{c}{Fixed} & 
\multicolumn{1}{c}{Moving} &
\multicolumn{1}{c}{No. of Registration Pairs} \\ \midrule

\multicolumn{1}{l}{$\mathbf{T1 \rightarrow T2}^{\text{a}}$} & 
\multicolumn{1}{c}{OASIS-3} & 
\multicolumn{1}{c}{OASIS-3} &
\multicolumn{1}{c}{100} \\

\multicolumn{1}{l}{$\mathbf{T1 \rightarrow T2}^{\text{b}}$} & 
\multicolumn{1}{c}{OASIS-3} & 
\multicolumn{1}{c}{IXI} &
\multicolumn{1}{c}{100} \\

\multicolumn{1}{l}{$\mathbf{T1 \rightarrow T2}^{\text{c}}$} & 
\multicolumn{1}{c}{IXI} & 
\multicolumn{1}{c}{IXI} &
\multicolumn{1}{c}{100} \\ \bottomrule
\end{tabular}
\end{table}

\subsubsection{Dataset selection}

For each of our experiments, the test sets are selected according to demographic priors based on their available attributes. In \textsl{IXI}, each scan belongs to a different healthy patient, so we randomly sample different patients and separate them into fixed or moving candidates to form registration pairs, regardless of the hospital where each sequence was collected. \textsl{OASIS-3} is longitudinal and the scans are sampled so that patients in fixed and moving test sizes do not coincide with each other in different periods of time. Fixed images to evaluate registration are sampled in a stratified manner: we divide the samples into healthy, suspected dementia, mild cognitive impairment, dementia, and severe dementia scans. This classification is addressed by the Clinical Dementia Rating Scale (CDR) available as demographic metadata for each subject and calculated in a personal interview evaluating cognitive abilities such as memory, orientation, or judgment.

\begin{figure}[t]
    \centering
    \includegraphics[width=0.5\textwidth]{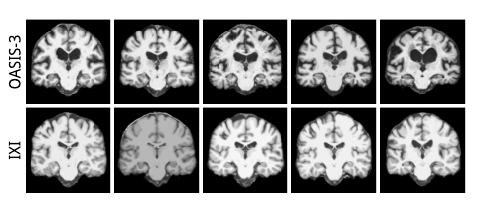}
    \caption{\textbf{Selected fixed images for our experiments}. We show the slices on the coronal plane of all the chosen fixed scans of all our evaluation pairs.}
    \label{fig:atlases-coronal-views}
\end{figure}

As each of the used datasets has different population properties, we augment our different experiments by different combinations of fixed or moving sequences. With this approach, we can test the performance of registration methods on large or small deformations. The corresponding experiments are related in \tref{table:test-registration-sets} and the coronal views of the chosen fixed images are shown in \fref{fig:atlases-coronal-views}. In this setup, we have considered OASIS-3 as the fixed images dataset in two of the experiments. The remaining benchmark contains only IXI samples to evaluate registration on small deformations in healthy subjects. To ensure proper reproducibility in further experiments of the tested methods, we detail the identifiers of the chosen fixed images in our experiments in \aref{appendix:fixed-images}.

\subsection{Baselines}
\label{subsec:baselines}

Our selected representatives of classical registration methods have been \texttt{ANTs} (\texttt{SyN}) \cite{avants2008syn}, \texttt{NiftyReg}, and \texttt{deedsBCV}. \texttt{ANTs} is run with four multiresolution levels and mutual information minimization between samples with different contrasts. \texttt{NiftyReg} \cite{modat2010niftyreg} registers with 10mm spacing in control points to obtain regularized solutions with normalized mutual information. Lastly, \texttt{deedsBCV} algorithm \cite{heinrich2013mrfdeeds} computes \texttt{MIND} features on a multiresolution strategy. Details on parameter configurations are provided in \aref{appendix:hyperparameter-values}.

Regarding learning-based baselines, we consider different state-of-the-art methods specialized on multimodal registration. These are classified into general-purpose and instance-specific registration methods. In the first category, we choose \texttt{OTMorph} \cite{kim2025otmorph}, trained for 3000 epochs with default parameters, \texttt{SynthMorph} \cite{hoffmann2022synthmorph} without fine-tuning and default regularization, and \texttt{TransMatch} \cite{chen2024transmatch}, trained with the Dice metric. In the second category,  we benchmark our datasets in \texttt{MultiGradICON} \cite{demir2024multigradicon, tian2023gradicon},  \texttt{ConvexAdam} \cite{heinrich2014convex, siebert2025convexadam} with \texttt{MIND} features and consistency regularization \cite{christensen2001consistentimageregistration}, and \texttt{DINO-Reg} \cite{song2024dinoreg}, which is based on the same proximal splitting solver as \texttt{ConvexAdam}, but uses self-supervised visual and semantic features from the \texttt{DINOv2} \cite{oquab2024dinov2} backbone \texttt{ViT-L/14}. This method extracts patch-level features with this \texttt{DINOv2} \cite{oquab2024dinov2} backbone and runs Principal Component Analysis (PCA) to reduce the dimensionality of the features. We run \texttt{DINO-Reg} registration by masking features in the whole brain anatomy after dimensionality reduction, in order to avoid variance on background.

From the original parameter configurations, we have changed the default configurations to improve the results in our brain MRI experiments. In particular, we have tuned instance-specific iterations in \texttt{MultiGradICON} to avoid potential data leakage between training and test pairs, and parameter values in \texttt{deedsBCV}, obtaining better results in our test sets than with defaults. We also adjusted the size of the final features in \texttt{DINO-Reg} as discussed in \aref{appendix:hyperparameter-values}. Except \texttt{SynthMorph} \cite{hoffmann2022synthmorph} and \texttt{TransMatch} \cite{chen2024transmatch}, the remaining baselines are trained without segmentation information. 

\subsection{Evaluation}
\label{subsec:evaluation}

We evaluate accuracy and smoothness by the Dice similarity coefficient (DSC) and the ratio of voxels with a negative Jacobian determinant $\det(J_{\varphi}) \leq 0$. The Dice score is obtained by the average of 32 anatomical structures predicted by \texttt{SynthSeg} \cite{billot2023synthseg}. The choice of \texttt{SynthSeg} as the evaluation segmentation engine was motivated by its robustness to varying MRI contrasts and image resolution. This segmentation engine is learning-based and registration pairs can potentially coincide with training from this segmentation network. However, unless specifically indicated, segmentation maps are used only for the evaluation of registration accuracy. These 32 anatomical labels include a set of representative brain structures:  Cerebral White Matter, Cortex and Lateral Ventricles, Cerebellar White Matter, Brain Stem or Hippocampus, among others.

\subsection{Experiment 1: Comparison with Multimodal Baselines}

We compare the proposed methods with other multimodal registration approaches for different sources of medical images. Some of these methods have been extensively evaluated on CT-to-MRI registration for abdominal sequences \cite{kim2025otmorph, song2025dinoreg}, for lung CT registration \cite{demir2024multigradicon, heinrich2013mrfdeeds} or exclusively for neuroimages \cite{yifan2022nodeo, hernandez2025pdelddmmnodeo, siebert2025convexadam, modat2010niftyreg, avants2008ants}.  Our setup consists of 100 data samples corresponding to different registration pairs for each of the chosen methods, in each of the experiments of \tref{table:test-registration-sets}. 

\begin{figure*}[t]
    \centering
    \includegraphics[width=0.8\linewidth]{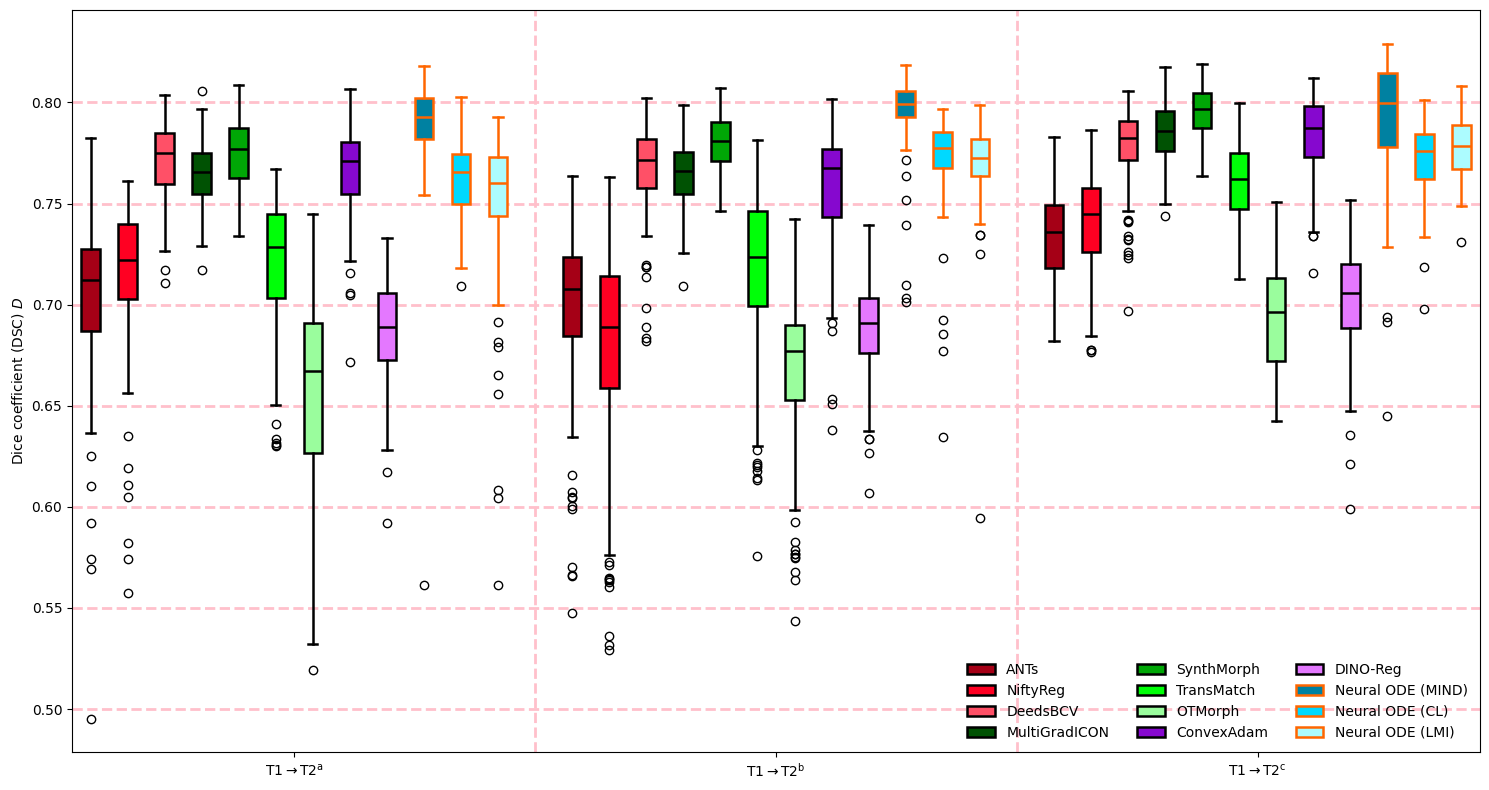}
    \caption{\textbf{Registration accuracy results}. We represent the distributions of the registration accuracy, measured by the Dice score coefficient (DSC). Orange-edges boxes correspond to our method proposals for each of our experiments in \tref{table:test-registration-sets}.}
    \label{fig:dsc-h95-main-baselines}
\end{figure*}

\begin{figure}[t]
    \centering
    \includegraphics[width=0.9\linewidth]{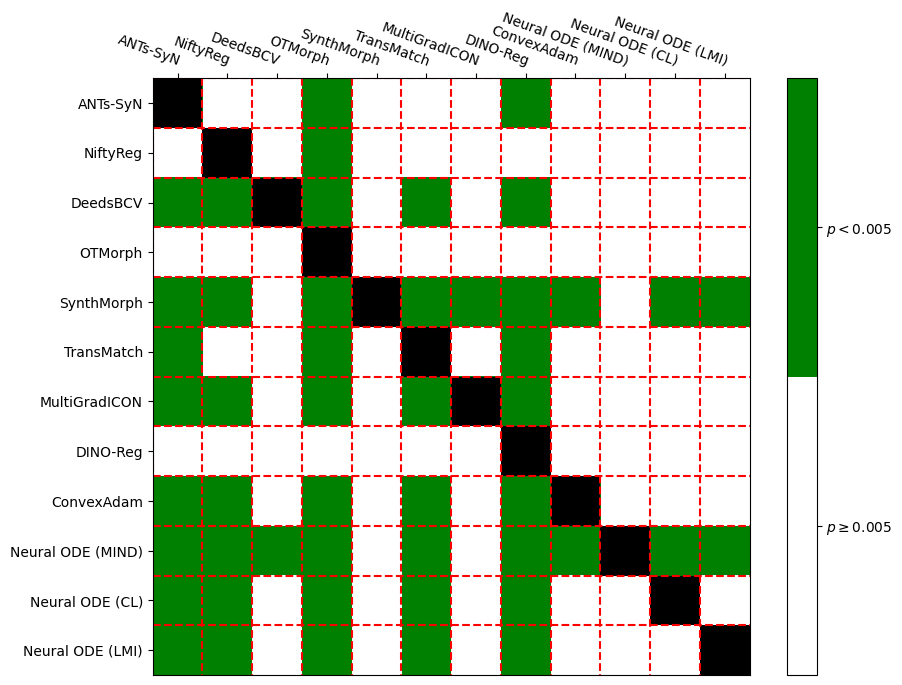}
    \caption{\textbf{Significance tests on our baseline comparisons}. We represent the significance results of the Wilcoxon tests on Dice distributions. The test run is one-sided and row methods are compared respect to column methods. We compute the p-values $p$ for each experiment $\text{T1} \rightarrow \text{T2}^{\text{a}}$, $\text{T1} \rightarrow \text{T2}^{\text{b}}$, $\text{T1} \rightarrow \text{T2}^{\text{c}}$, namely $p_{a}$, $p_{b}$ and $p_{c}$, and consider statistical significance for model comparison if $\max\{p_{a}, p_{b}, p_{c}\} < 0.005$. Black labels in the p-value matrix correspond to trivial comparisons in the diagonal.}
    \label{fig:pvalues-matrix-baselines}
\end{figure}

We tested the statistical differences of the average Dice scores on the 32 segmented anatomical labels. We use a right-sided nonparametric Wilcoxon signed-rank test \cite{woolson2005wilcoxonsignedrank} to assess whether some chosen Dice distribution is greater than any of the benchmark methods in this work. An overview of the complete distribution of the mean Dice evaluation of the different methods for each of our experiments is shown in \fref{fig:dsc-h95-main-baselines} and the significance tests in \fref{fig:pvalues-matrix-baselines}. For each pair of methods to compare, we compute a p-value $p$ on each experiment of \tref{table:test-registration-sets} and conclude the significant difference if $\max_{i} \{ p_{i} \} \leq 0.005$, considering the maximum over the different experiment families at \tref{table:test-registration-sets}. In these comparisons, we indicate the maximum of these p-values as an upper bound.

We begin our comparison tests with classical registration methods. We distinguish between general-purpose methods optimized with normalized mutual information (\texttt{ANTs} or \texttt{NiftyReg}) and feature-based deformable registration (\texttt{DeedsBCV}). \texttt{ANTs} registration is outperformed by our methods using structural or nonstructural metric models in all our experiments, either \texttt{MIND} ($p \leq 3.2 \cdot 10^{-15}$) or contrastive descriptors ($p \leq 2.1 \cdot 10^{-16}$) and local mutual information ($p < 6.4 \cdot 10^{-12}$). By analogous analysis, \texttt{NiftyReg} is outperformed using either \texttt{MIND} ($p \leq 3.0 \cdot 10^{-15}$), contrastive learning ($p \leq 2.9 \cdot 10^{-14}$) or local mutual information ($p \leq 6.8 \cdot 10^{-11}$) together with the Neural ODE backbone. Compared to \texttt{DeedsBCV}, our method performs poorly when trained with contrastive descriptors or local mutual information, and our tests do not give any significance ($p \gg 0.005$) for any of our experiments, although it still outperforms when registering with \texttt{MIND} features ($p \leq 1.5 \cdot 10^{-5}$). 

With respect to learning-based methods, we train \texttt{TransMatch} and \texttt{OTMorph} from scratch with our training partitions on each of the related experiments and use the foundational models available for evaluation of \texttt{MultiGradICON} and \texttt{SynthMorph}. In the last two cases, both models have been originally trained with each of the modalities used in this work, \texttt{MultiGradICON} has been fine-tuned on each registration pair with 100 iterations and \texttt{SynthMorph} remains in its foundational state. Importantly, our method with \texttt{MIND} features outperforms \texttt{SynthMorph} in the $\text{T1} \rightarrow \text{T2}^{\text{a}}$ ($p = 1.5 \cdot 10^{-9} \leq 0.005$) and $\text{T1} \rightarrow \text{T2}^{\text{b}}$ ($p = 3.1 \cdot 10^{-12} \leq 0.005$) experiments and \texttt{MultiGradICON} in all of them ($p < 0.002$) in the same experiments ($p = 4.2 \cdot 10^{-14}$ and $p = 6.5 \cdot 10^{-16} \leq 0.005$ for $\text{T1} \rightarrow \text{T2}^{\text{a}}$ and $\text{T1} \rightarrow \text{T2}^{\text{b}}$, respectively). There is no evidence to ensure better performance for contrastive descriptors or local mutual information with Neural ODEs and these methods, even though all these variants outperform \texttt{TransMatch} and \texttt{OTMorph} ($p < 1 \cdot 10^{-5}$ and $p < 1.6 \cdot 10^{-16}$, respectively).

Similarly, as in \texttt{SynthMorph}, it is still relevant that in experiment $\text{T1} \rightarrow \text{T2}^{c}$ there is weaker evidence of outperformance with \texttt{ConvexAdam} than in the remaining experiments for the Neural ODE with \texttt{MIND} features ($p \simeq 0.002$). For localized mutual information and contrastive descriptors, both \texttt{SynthMorph} and \texttt{ConvexAdam} outperform. The superiority of \texttt{MIND} features for modality-agnostic registration is demonstrated in other possible comparisons in our benchmarks, since methods like \texttt{DINO-Reg} share the same proximal splitting method as \texttt{ConvexAdam}, but significantly underperform in our setting ($p \ll 0.005$ compared to \texttt{ANTs}) for registration in neuroimages. 

\begin{table}[t]
        \caption{Ratios (\%) of negative Jacobians for each of our benchmark experiments.}
        \label{table:ratios-negative-jacobians-determ}
        \scalebox{0.85}{
        \begin{tabular}{@{}llll@{}}
        \toprule
        \multicolumn{1}{l}{\textbf{Method}} & 
        \multicolumn{3}{c}{\% $J_{\varphi} \leq 0$ $\, \downarrow$} \\ \midrule
    
        \multicolumn{1}{l}{\textbf{Experiment}} & 
        \multicolumn{1}{l}{$\textbf{T1} \rightarrow \textbf{T2}^{\text{a}}$} & 
        \multicolumn{1}{l}{$\textbf{T1} \rightarrow \textbf{T2}^{\text{b}}$} &  \multicolumn{1}{l}{$\textbf{T1} \rightarrow \textbf{T2}^{\text{c}}$} \\ \midrule
        
        \multicolumn{1}{l|}{ANTs-SyN \cite{avants2008syn}} & 
        \multicolumn{1}{c}{\cellcolor{mygreen}$0$} & 
        \multicolumn{1}{c}{\cellcolor{mygreen}$0$} & 
        \multicolumn{1}{c}{\cellcolor{mygreen}$0$} \\
    
        \multicolumn{1}{l|}{NiftyReg \cite{modat2010niftyreg}} & 
        \multicolumn{1}{c}{$9.72 \cdot 10^{-3}$} &
        \multicolumn{1}{c}{$2.78 \cdot 10^{-3}$} &
        \multicolumn{1}{c}{$2.55 \cdot 10^{-3}$} \\

        \multicolumn{1}{l|}{DeedsBCV \cite{heinrich2013mrfdeeds, heinrich2015multimodaldeedsbcv}} & 
        \multicolumn{1}{c}{$1.84 \cdot 10^{-1}$} &
        \multicolumn{1}{c}{$1.75 \cdot 10^{-1}$} &
        \multicolumn{1}{c}{$1.99 \cdot 10^{-1}$} \\ \midrule

        \multicolumn{1}{l|}{OTMorph \cite{kim2025otmorph}} & 
        \multicolumn{1}{c}{$6.41 \cdot 10^{-3}$} &
        \multicolumn{1}{c}{$1.43 \cdot 10^{-4}$} &
        \multicolumn{1}{c}{$9.89 \cdot 10^{-6}$} \\

        \multicolumn{1}{l|}{SynthMorph \cite{hoffmann2022synthmorph}} &    
        \multicolumn{1}{c}{\cellcolor{mygreen}$0$} & 
        \multicolumn{1}{c}{\cellcolor{mygreen}$0$} & 
        \multicolumn{1}{c}{\cellcolor{mygreen}$0$} \\
        
        \multicolumn{1}{l|}{TransMatch \cite{chen2024transmatch}} & 
        \multicolumn{1}{c}{$2.21 \cdot 10^{-3}$} &
        \multicolumn{1}{c}{$8.25 \cdot 10^{-4}$} &
        \multicolumn{1}{c}{$1.67 \cdot 10^{-4}$} \\

        \multicolumn{1}{l|}{MultiGradICON (\texttt{MIND}) \cite{demir2024multigradicon}} & 
        \multicolumn{1}{c}{$2.69 \cdot 10^{-2}$} & 
        \multicolumn{1}{c}{$3.02 \cdot 10^{-2}$} & 
        \multicolumn{1}{c}{$1.67 \cdot 10^{-2}$} \\ 

        \multicolumn{1}{l|}{DINO-Reg \cite{oquab2024dinov2, song2025dinoreg}} & 
        \multicolumn{1}{c}{$3.11 \cdot 10^{-2}$} &
        \multicolumn{1}{c}{$3.25 \cdot 10^{-2}$} &
        \multicolumn{1}{c}{$3.31 \cdot 10^{-2}$} \\

        \multicolumn{1}{l|}{ConvexAdam (\texttt{MIND}) \cite{heinrich2014convex, siebert2025convexadam}} & 
        \multicolumn{1}{c}{$1.61 \cdot 10^{-2}$} &
        \multicolumn{1}{c}{$2.29 \cdot 10^{-2}$} &
        \multicolumn{1}{c}{$1.64 \cdot 10^{-2}$} \\ \midrule
        
        \multicolumn{1}{l|}{Neural ODE (\texttt{MIND})} & 
        \multicolumn{1}{c}{$3.97 \cdot 10^{-2}$} &
        \multicolumn{1}{c}{$5.08 \cdot 10^{-2}$} &
        \multicolumn{1}{c}{$3.85 \cdot 10^{-2}$} \\
        
        \multicolumn{1}{l|}{Neural ODE (\texttt{CL})} & 
        \multicolumn{1}{c}{$4.64 \cdot 10^{-2}$} &
        \multicolumn{1}{c}{$6.06 \cdot 10^{-2}$} &
        \multicolumn{1}{c}{$5.06 \cdot 10^{-2}$} \\

        \multicolumn{1}{l|}{Neural ODE (\texttt{LMI})} & 
        \multicolumn{1}{c}{$1.46 \cdot 10^{-1}$} &
        \multicolumn{1}{c}{$1.70 \cdot 10^{-1}$} &
        \multicolumn{1}{c}{$9.48 \cdot 10^{-2}$} \\ \midrule
        \end{tabular}
    }
\end{table}

\subsection{Experiment 2: Registration with Learned Descriptors}

One of the structural metrics proposed in this work is a differentiable approximation by the neural network $F_{\theta}$ from \sref{sec:differentiable-approximation-contrastive}. This approach learns a structural descriptor by contrastive learning, so $\mathcal{S}$ from \eref{eq:variational-registration} becomes the normalized distance between the anatomically corresponding descriptors computed in feature space \cite{ronchetti2023disa, mok2024modalityagnostic, pielawski2020comir}.

\begin{figure*}[t]
    \centering
    \includegraphics[width=0.9\linewidth]{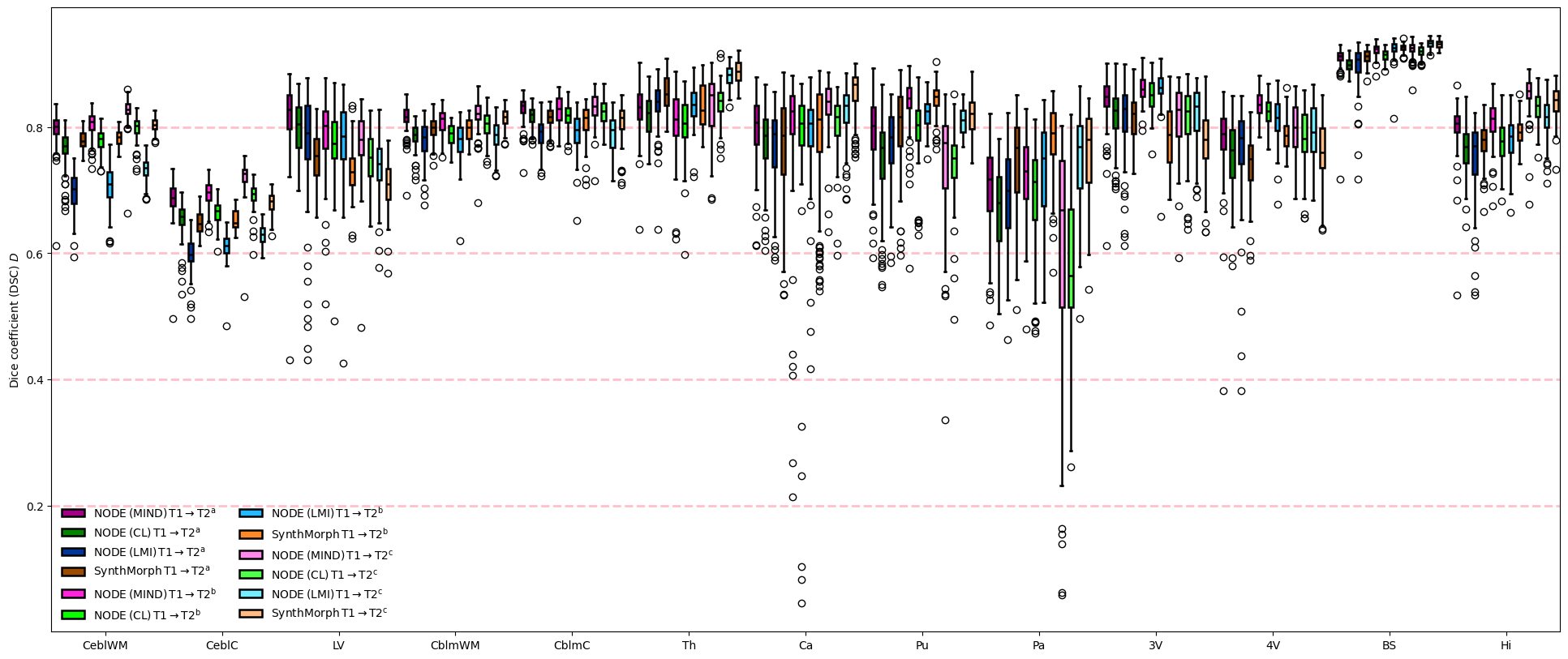}
    \caption{\textbf{Fine-grain evaluation on multimodal baselines}. We measure Dice score coefficient on different anatomical regions of interest. These regions are, from left to right: Cerebral White Matter (CeblWM), Cerebral Cortex (CeblC), lateral ventricles (LV), Cerebellum White Matter (CblmWM), Thalamus (Th), Caudate (Ca), Putamen (Pu), Pallidum (Pa), 3rd Ventricle (3V), 4th Ventricle (4V), Hippocampus (Hi) and Brain Stem (BS).}
    \label{fig:fine-grain-dsc}
\end{figure*}

We perform a fine-grain evaluation of registration accuracy on 13 anatomical structures in \fref{fig:fine-grain-dsc} in all of our experiments, among Neural ODE methods with structural metrics, local mutual information, and \texttt{SynthMorph}. As a relevant effect, conditional distributions to certain anatomical regions, like the Pallidum (Pa), have higher variance (and hence, variability) than others which comprehend more volume, like the Cerebral Cortex (CblC) when we address registration with contrastive descriptors. 

\begin{figure}[t]
    \centering
    \includegraphics[width=0.5\textwidth]{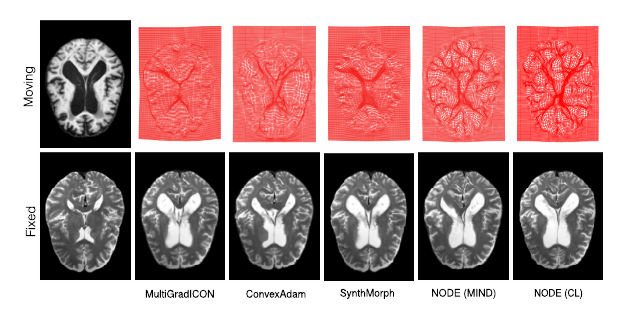}
    \caption{\textbf{Qualitative comparison of registration example}. We choose a challenging registration pair with atrophy in the ventricles. The proposed approaches register qualitatively most of the structures and their accuracy is measured by mean of 32 structures at different finer or coarser scales.}
    \label{fig:qualitative-registration-example}
\end{figure}

There is no strong response on test distributions to consider outperformance in certain structures such as the Cerebral cortex when comparing distributions between \texttt{MIND}-based registration over the contrastive learning approach. However, these structural descriptors are formed by two dilation factors in the feature space and show qualitative outperformance in registering large-scale differences compared to other state-of-the-art methods, as \fref{fig:qualitative-registration-example} depicts. The evaluation by mean Dice score obscures this fact in the complete baseline comparison for our paired comparisons.

\subsection{Experiment 3: Penalty-barrier Regularization Analyses}

Our registration framework learns velocity fields $v_{\theta}$ by a neural network conditioned on the $V-$regularizer from \eref{eq:node-optimization-target}. As a relevant result, the structure of each chosen neural network encodes prior knowledge about low-level image features for different energy-minimization tasks \cite{wu2024nodesequential, ulyanov2018deepprior}. Regarding our method, it uses a convolution architecture together with linear layers to encode pointwise displacements, given its proven efficiency for monomodal registration \cite{hernandez2025pdelddmmnodeo, yifan2022nodeo}.

The learned velocity fields $v_{\theta}$ are trained by joint optimization of our similarity functions with different regularizers in \eref{eq:node-optimization-target} to encourage local diffeomorphic properties in $\varphi$. In particular, regularization for orientation-preserving solutions $\varphi$ is accomplished by the parameter $\varepsilon$ in $\mathcal{L}_{J}$ from \eref{eq:node-optimization-target}. In our experiments in \fref{fig:dsc-h95-main-baselines}, we empirically set a fixed value for $\varepsilon$ to balance the tradeoff between registration accuracy and ratios of negative Jacobians given this flexibility. The resulting negative Jacobian ratios are similar in order to other baseline methods and are related in \tref{table:ratios-negative-jacobians-determ}.

\begin{figure*}[t]
    \centering
    \def\svgwidth{\linewidth}
    \includegraphics[width=\textwidth]{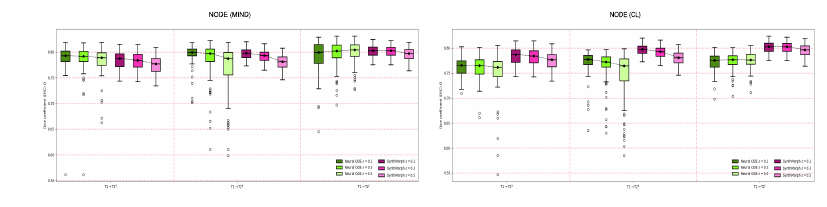}
    \scalebox{0.9}{
    \begin{tabular}{@{}llllllllll@{}}
    \toprule
    \multicolumn{1}{l}{\textbf{Method}} & 
    \multicolumn{9}{c}{\% $J_{\varphi} \leq 0$ $\, \downarrow$} \\ \midrule
    \multicolumn{1}{l|}{\textbf{Experiment}} & 
    \multicolumn{3}{c|}{$\textbf{T1} \rightarrow \textbf{T2}^{\text{a}}$} & 
    \multicolumn{3}{|c|}{$\textbf{T1} \rightarrow \textbf{T2}^{\text{b}}$} &  
    \multicolumn{3}{|c}{$\textbf{T1} \rightarrow \textbf{T2}^{\text{c}}$} \\ \midrule

    \multicolumn{1}{l|}{\text{\textbf{Penalty-barrier} $\varepsilon$}} & 
    \multicolumn{1}{c}{$\varepsilon = 0.1$} & 
    \multicolumn{1}{c}{$\varepsilon = 0.3$} & 
    \multicolumn{1}{c|}{$\varepsilon = 0.5$} & 
    \multicolumn{1}{c}{$\varepsilon = 0.1$} & 
    \multicolumn{1}{c}{$\varepsilon = 0.3$} & 
    \multicolumn{1}{c|}{$\varepsilon = 0.5$} & 
    \multicolumn{1}{c}{$\varepsilon = 0.1$} & 
    \multicolumn{1}{c}{$\varepsilon = 0.3$} & 
    \multicolumn{1}{c}{$\varepsilon = 0.5$} \\ \midrule
    
    \multicolumn{1}{l|}{Neural ODE (\texttt{MIND})} & 
    \multicolumn{1}{c}{$3.97 \cdot 10^{-2}$} & 
    \multicolumn{1}{c}{$4.81 \cdot 10^{-5}$} & 
    \multicolumn{1}{c|}{\cellcolor{mygreen}$0$} &
    \multicolumn{1}{c}{$5.08 \cdot 10^{-2}$} & 
    \multicolumn{1}{c}{$8.58 \cdot 10^{-5}$} & 
    \multicolumn{1}{c|}{\cellcolor{mygreen}$0$} &
    \multicolumn{1}{c}{$3.85 \cdot 10^{-2}$} & 
    \multicolumn{1}{c}{$3.89 \cdot 10^{-5}$} & 
    \multicolumn{1}{c}{\cellcolor{mygreen}$0$} \\

    \multicolumn{1}{l|}{Neural ODE (\texttt{CL})} & 
    \multicolumn{1}{c}{$4.64 \cdot 10^{-2}$} &
    \multicolumn{1}{c}{$4.02 \cdot 10^{-4}$} &
    \multicolumn{1}{c|}{\cellcolor{mygreen}$0$} &
    \multicolumn{1}{c}{$6.06 \cdot 10^{-2}$} &
    \multicolumn{1}{c}{$2.38 \cdot 10^{-4}$} &
    \multicolumn{1}{c|}{\cellcolor{mygreen}$0$} &
    \multicolumn{1}{c}{$5.06 \cdot 10^{-2}$} &
    \multicolumn{1}{c}{$1.87 \cdot 10^{-4}$} &
    \multicolumn{1}{c}{\cellcolor{mygreen}$0$} \\

    \multicolumn{1}{l|}{SynthMorph} & 
    \multicolumn{1}{c}{\cellcolor{mygreen}$0$} &
    \multicolumn{1}{c}{\cellcolor{mygreen}$0$} &
    \multicolumn{1}{c|}{\cellcolor{mygreen}$0$} &
    \multicolumn{1}{c}{\cellcolor{mygreen}$0$} &
    \multicolumn{1}{c}{\cellcolor{mygreen}$0$} &
    \multicolumn{1}{c|}{\cellcolor{mygreen}$0$} &
    \multicolumn{1}{c}{\cellcolor{mygreen}$0$} &
    \multicolumn{1}{c}{\cellcolor{mygreen}$0$} &
    \multicolumn{1}{c}{\cellcolor{mygreen}$0$} \\ 
    \bottomrule
    \end{tabular}
    }
    \caption{\textbf{Regularization analysis}. For varying values of the parameter $\varepsilon$, we study the performance of our method in terms of Dice overlap and Jacobian determinants in comparison with \texttt{SynthMorph}. We perform this evaluation in both the Neural ODE backbone with \texttt{MIND} and contrastive features in the first and second rows respectively.}
    \label{fig:boxplots-dsc-rnj-node-eps}
\end{figure*}

We analyze the effect of varying the parameter $\varepsilon$ of the penalty term $\mathcal{L}_{J}$ from \eref{eq:node-optimization-target} by sampling values ${\varepsilon \in \{ 0.1, 0.3, 0.5\}}$ and measure the quality of the resulting models in \fref{fig:boxplots-dsc-rnj-node-eps} for our variants with structural descriptors. Increasing the values in $\varepsilon$ imposes a harder constraint on the minimum value of the Jacobian determinants if they are negative. In all cases, Jacobian ratios are computed throughout the image domain, i.e. $160 \cdot 224 \cdot 192 \simeq 6.8 \cdot 10^{6}$ voxels. We sample the same values on the regularization strength for \texttt{SynthMorph} and compare the registration results with our method.

\begin{figure*}[t]
    \centering
    \includegraphics[width=\linewidth]{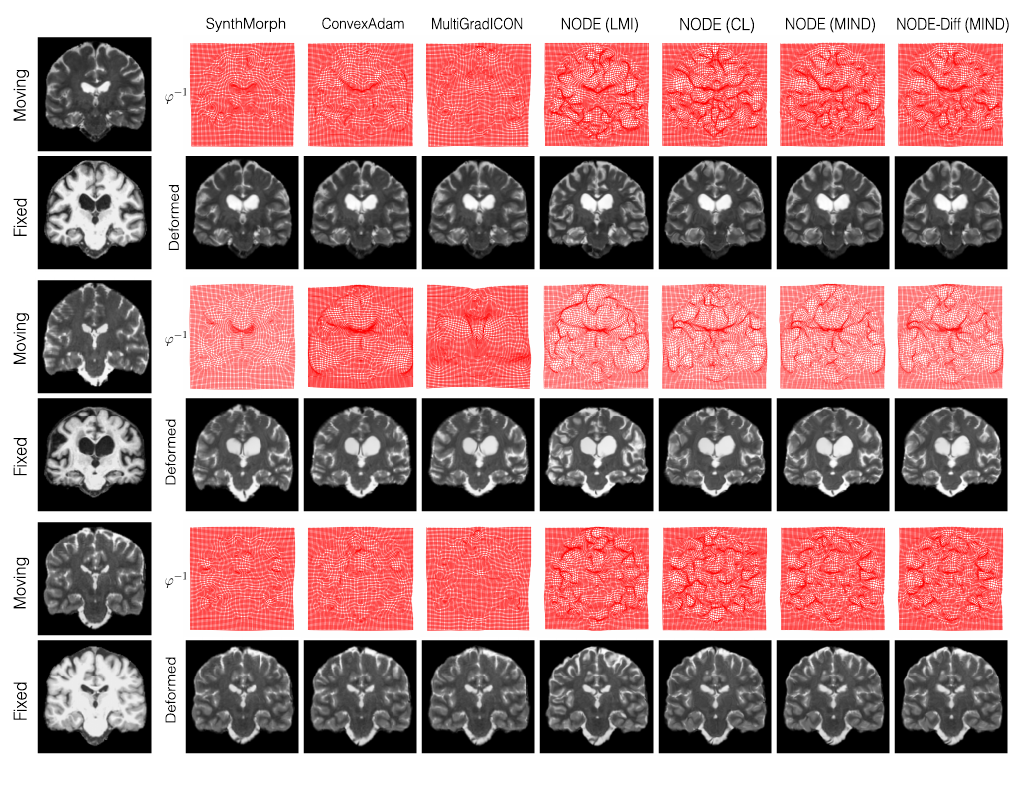}
    \caption{\textbf{Qualitative examples}. We show qualitative results of the registration results for three samples of each of our experiments $\text{T1} \rightarrow \text{T2}^{\text{a}}$, $\text{T1} \rightarrow \text{T2}^{\text{c}}$, $\text{T1} \rightarrow \text{T2}^{\text{c}}$, ordered from top to bottom, in the best performing baselines chosen for this work.}
    \label{fig:synth-node-qualitative-results}
\end{figure*}

The resulting model variants on different values of $\varepsilon$ bring to the conclusion that our method is able to regularize locally while not drastically penalizing the performance in our datasets, as  seen in qualitative examples like \fref{fig:synth-node-qualitative-results}, where we included a variant \texttt{NODE-Diff} with maximum regularization corresponding to $\varepsilon = 0.5$ from \fref{fig:boxplots-dsc-rnj-node-eps}. These experiments also confirm that the tradeoff in regularity and accuracy is not linear, as in experiment $\text{T1} \rightarrow \text{T2}^{\text{c}}$ the resulting medians outperform \texttt{SynthMorph} mark in terms of median. This result is indeed related to the definition of the experiment $\text{T1} \rightarrow \text{T2}^{\text{c}}$, as all samples are healthy and small deformation methods are sufficient in this setting.

Possible extensions of regularization analysis would contemplate other schemes, such as spatially-varying weights \cite{niethammer2019metriclearning, chen2023spatiallyvarying} or other smoothness regularizers based on seminorms \cite{mang2016h1regularizers}. We restrict ourselves to this grid search on penalty-barrier hyperparameters for the evaluation of the regularization limits of our approach.
\section{Discussion}

The clinical relevance of our multimodal registration approach lies in its ability to provide plausible deformations in the presence of modality shifts and large structural variation. By alleviating the registration under-performance that typically arises when aligning different MRI modalities, our method has the potential to substantially improve the reliability of Computational Anatomy studies.  This is fundamentally useful in the study of neurodegenerative diseases, where inter-subject paired modalities are often scarce and subtle anatomical changes must be detected with high fidelity.

Our approach achieves high registration accuracy in challenging shape variability, mainly determined by different degrees of anatomical atrophy, and is therefore adequate for evaluation in large deformations. Given the ill-posedness of registration in this paradigm, we discuss the influence of the key components of these registration variants on its final output.

\subsection{Time Complexity}

Our experiments were performed on an NVIDIA GeForce RTX 5090. In this setup, the Neural ODE backbone performs registration at a mean time of $67.01 \pm 0.89$s  per registration pair with the \texttt{MIND} descriptor. This method consists solely of the Neural ODE network with a fixed step size and a one step solver computed by Euler method, so it performs two neural network evaluations for each training iteration: one for the forward pass and another for the gradient computation by integrating the adjoint. The contrastive token is computed with the frozen descriptor network, but the Neural ODE optimization needs to perform inference at each iteration of the descriptor of the warped image, and the registration time increments to $130.56 \pm 0.30$s. This time difference does not result in a more accurate registration result or more regular solutions, as we have seen in \fref{fig:dsc-h95-main-baselines}, and \texttt{MIND} features outperform in our setting against contrastive learning given the registration accuracy results at \fref{fig:dsc-h95-main-baselines}. 

Regarding other instance-specific methods, \texttt{MultiGradICON} finishes after 100 instance-specific epochs by fine tuning from its foundational weights with average time $67.08 \pm 0.40$s, \texttt{ConvexAdam} by $7.53 \pm 0.49$s or \texttt{DINO-Reg} $208.59 \pm 0.92$s, slowed down by linear resizing of feature maps and inference on the \texttt{DINOv2} backbone. Learning-based methods register fast at inference time and their training time depends on each current architecture.

The best performance is achieved with methods that run in around $65$s per registration pair. The most advantageous algorithm in terms of efficiency is clearly \texttt{ConvexAdam}, since it does not train a deep network and instead minimizes a functional in the displacement of source features \cite{steinbrucker2009largedeformation} together with a composition consistency regularizer \cite{christensen2001consistentimageregistration}, given that its results perform better than most of the baselines. For our experiments, we claim that a large deformation method was more suitable to evaluate inter-subject registration between sequences of healthy or non-healthy patients.

\subsection{Regularization Sensitivity}

We implicitly consider that the estimated transformation $\varphi$ is diffeomorphic for registration evaluation if it satisfies local invertibility and preserves orientation by the condition $\det (\text{Jac}[\varphi](\mathbf{x})) > 0$ for all $\mathbf{x} \in \Omega$ in the image domain. However, it is well-known that optimizing $v$ by penalty regularizers on $\det(\text{Jac}[\varphi])$ does not avoid injectivity violations on the transformed domain $\varphi(\Omega)$. This has been argued in seminal work like \cite{burguer2013hyperelasticregularizer}, so it is necessary to geometrically constrain the domain $\varphi(\Omega)$.

Our method maintains performance even by imposing harder constraints on $\det(\text{Jac}[\varphi])$ to be strictly positive.  To this extent, our proposed approach constitutes a reasonable method that balances the existing tradeoff between accuracy, regularization on efficient optimization for current network-based architectures, and time efficiency. Although we consider our method to be suitable for general deformable registration, registration accuracy can be degraded in fine-grain anatomical evaluation with a highly-regularized setting, as \fref{fig:synth-node-qualitative-results} depicts, but average performance is only subtly degraded by anti-folding (Jacobian) regularization.

\subsection{Spatial Search}

The patch search region of structural descriptors is parameterized by the parameters of the receptive field $r$ and the dilation factor $d_{r}$ of \sref{subsec:structural-descriptors}. Choosing different values of these two parameters conditions the resulting features to be more discriminative on coarser or finer scales, and we select default values for these two parameters given previous studies on the best performing alternatives on average \cite{siebert2025convexadam} and by hyperparameter tuning in our experiments.

In this work, the descriptors are computed with two different search schemes. Regarding the \texttt{MIND} descriptor, we chose $d_{r} = 2$ and $r = 1$ throughout this work. Taking into account the contrastive learning descriptors, we considered $d_{r} \in \{1, 2\}$ and stacked these two descriptors to form a token of double length than \texttt{MIND} features like in \cite{mok2024modalityagnostic}. However, although  qualitative results show better registration on coarser scales, as in the case of \fref{fig:qualitative-registration-example}, mean Dice scores do not give any significance of the superiority of this method over \texttt{MIND} features. With this observation, the selected parameters best leverage the tradeoff of finer or coarser registration on computed self-similarities for descriptor-based variants.
\section{Conclusions}

Our work establishes and validates the potential of local self-similarity descriptors in the context of multimodal registration with learning methods for pairwise registration. This framework  works exclusively in image space, rapidly achieves accurate registration, and is robust to hard-constrained explicit regularization. This work provides a solid proof on these  properties for large deformation registration, and we expect this is a promising direction to perform extensive evaluation on further applications with different domain shifts.

Regarding future directions in this work, we highlight two possible outcomes. In the same way some of our selected baselines were originally tested in different modalities on different anatomies than neuro-anatomy, we believe that the superiority of the \texttt{MIND} features will be maintained compared to other structural descriptors. Lastly, our framework may be extended for other applications requiring multimodal registration, for example dynamic characterization or longitudinal registration for large motion with periodic boundary conditions \cite{wu2024nodesequential}. Such applications are fundamental in spatiotemporal methods and permit rich usability studies, such as atlas estimation across domains or motion modeling.

\section*{Acknowledgments}
IXI dataset is available by EPSRC GR/S21533/02  at \href{https://brain-development.org/ixi-dataset/}{brain-development.org}. The OASIS-3 dataset is provided by \textsl{OASIS-3: Longitudinal Multimodal Neuroimaging}. The authors thank Ubaldo Ramon-Julvez and Carlos Paesa-Lia for their proofreading and help in arranging the final figures. We also thank I3A from University of Zaragoza for their support in using their HPC cluster HERMES.

\appendix
\subsection{Fixed images}
\label{appendix:fixed-images}

\subsubsection{OASIS-3}
\texttt{OAS30088}, \texttt{OAS30273}, \texttt{OAS30884}, \texttt{OAS30110}, \texttt{OAS30682}. 

\subsubsection{IXI}
\texttt{IXI186}, \texttt{IXI027}, \texttt{IXI344}, \texttt{IXI560}, \texttt{IXI458}.

\subsection{Baseline parameters}
\label{appendix:hyperparameter-values}

Unless stated otherwise, we run baseline methods with default parameters. 

\subsubsection{DeedsBCV}

We considered a grid spacing of $6 \times 5 \times 4 \times 3 \times 2$, a search radius of $6 \times 5 \times 4 \times 3 \times 2$, and a quantization of the search step of $5 \times 4 \times 3 \times 2 \times 1$ where each axis refers to a multiresolution level in coarse-to-fine order.

\subsubsection{MultiGradICON}

We run 100 iterations of pairwise optimization on the foundational model weights, since we get better empirical results than the default 50 iterations.

\subsubsection{DINO-Reg}

We run 1000 iterations of the proximal splitting solver to compute the displacements in feature space. We use \texttt{DINOv2} backbone with $\texttt{ViT-L/14}$ weights, which has a patch size of 14. This method runs inference over all coronal slices sampled in intervals of 3 and computes the $80 \times 80$ features after bilinear upsampling. We use consistency regularization \cite{christensen2001consistentimageregistration} and low-rank PCA to reduce the feature components to a length of 64.


\section*{References}
\vspace{-1.5em}
\bibliography{egbib.bib}

\end{document}